\documentclass{llncs}
\usepackage{graphicx}

\usepackage{amsthm}
\usepackage{multirow}
\usepackage{url}
\usepackage{hyperref}

\begin{document}

\title{On Reasoning with RDF Statements about Statements using Singleton Property Triples}
\author{Vinh Nguyen\inst{1}, Olivier Bodenreider\inst{2},  Krishnaprasad Thirunarayan\inst{1}, Gang Fu\inst{3}, Evan Bolton\inst{3}, N\'{u}ria Queralt Rosinach\inst{4}, Laura I. Furlong\inst{4}, \\Michel Dumontier\inst{5}, Amit Sheth\inst{1} }
\institute{ \email{vinh,tkprasad,amit@knoesis.org}\\Kno.e.sis Center, Wright State University, Ohio, USA \and \email{olivier@nlm.nih.gov}\\National Library of Medicine, National Institute of Health, Maryland, USA \and \email{gang.fu@nih.gov}, \email{bolton@ncbi.nih.gov}\\ National Center of Biotechnology Information, National Institute of Health, \and \email{nqueralt,lfurlong@imim.es} \\Universitat Pompeu Fabra, Barcelona, Spain \and \email{michel.dumontier@stanford.edu}\\Stanford Center for Biomedical Informatics, Stanford University, California, USA}
\maketitle
\pagestyle{empty}

\begin{abstract}


The Singleton Property (SP) approach has been proposed for representing and querying metadata about RDF triples such as provenance, time, location, and evidence. In this approach, one singleton property is created to uniquely represent a relationship in a particular context, and in general, generates a large property hierarchy in the schema. It has become the subject of important questions from Semantic Web practitioners. Can an existing reasoner recognize the singleton property triples? And how? If the singleton property triples describe a data triple, then how can a reasoner infer this data triple from the singleton property triples? Or would the large property hierarchy affect the reasoners in some way? We address these questions in this paper and present our study about the reasoning aspects of the singleton properties. We propose a simple mechanism to enable existing reasoners to recognize the singleton property triples, as well as to infer the data triples described by the singleton property triples. We evaluate the effect of the singleton property triples in the reasoning processes by comparing the performance on RDF datasets with and without singleton properties. Our evaluation uses as benchmark the LUBM datasets and the LUBM-SP datasets derived from LUBM with temporal information added through singleton properties. 


\end{abstract}

\section{Introduction}

``The basic model contains just the concept of an assertion, and the concept of quotation - making assertions about assertions. This is introduced because (a) it will be needed later anyway and (b) most of the initial RDF applications are for data about data (\textit{metadata}) in which assertions about assertions are basic, even before logic." (Tim Berners-Lee, Semantic\footnote{http://www.w3.org/DesignIssues/Semantic.html}, 1998). Note that in Semantic Web, the term \textit{assertion} can be used interchangeably for \textit{triple} or \textit{statement}. 

The RDF data model contains two basic concepts: the concept of triples and the concept of triples about triples. An RDF triple $t$ consists of a subject $s$, a predicate $p$, and an object $o$. This concept of an RDF triple is intuitive and formally defined within the latest W3C Recommendations such as the RDF 1.1 concept and abstract syntax \cite{cyganiak2014}, and the RDF 1.1 formal semantics \cite{hayesrdf}. For triples about triples, RDF reification was previously presented as a standard in the W3C 2004 Recommendations \cite{hayes2004rdf}. However, this RDF reification has been withdrawn from the normative sections in the latest RDF Recommendation \cite{hayesrdf} and has become non-standard because of its limitations. Therefore, the concept of triples about triples has not yet been formally defined in the latest W3C RDF Recommendations. Representing triples about triples is a basic requirement for the RDF data model but remains an unresolved problem. 

To further motivate the need for a good mechanism to represent triples about triples and reason on them, we use the following running example. 

\textbf{Motivating example.}
\label{motivating}
We adopt the example from the popular Lehigh University Benchmark (LUBM) \cite{guo2005lubm}. Consider the \verb|worksFor| relationship between professors and universities: it does not represent the duration of the relationship, which is very common in a real world setting. We add the temporal dimension into this LUBM example as provided in Table \ref{motivating-example} (prefixes are eliminated for readability).

\begin{table}
\centering
\caption{Motivating example with temporal data to be represented in RDF\label{motivating-example}}
\begin{tabular}{l @{\hspace{1em}}l @{\hspace{1em}}l @{\hspace{1em}}l@{\hspace{1em}}l@{\hspace{1em}}l}  \hline\noalign{\smallskip}
Triple & Subject & Predicate & Object \\\hline \noalign{\smallskip}
$M_0$: & ProfessorA & worksFor & University1 \\\noalign{\smallskip} 
$M_1$: & $M_0$ & from & 1994\\\noalign{\smallskip} 
$M_2$: & $M_0$ & to & 2006\\\noalign{\smallskip} 
$M_3$: & ProfessorA & worksFor & University2 \\\noalign{\smallskip} 
$M_4$: & $M_3$ & from & 2007\\\noalign{\smallskip} 
$M_5$: & StudentB & hasAdvisor & ProfessorA \\\noalign{\smallskip} 
$M_6$: & $M_5$ & from & 2009\\\noalign{\smallskip} 
$M_7$: & worksFor & subPropertyOf & memberOf\\ \noalign{\smallskip}\hline\noalign{\smallskip}
\end{tabular}
\end{table}

Next, we discuss some kinds of metadata about the triples that are commonly used across various domains.

\textbf{Time and Space}. Knowledge evolves over time, and a proposition may only hold at a specific time point, duration, or interval \cite{hayes1996catalog}. An event may happen at one specific time and location. In the example at hand, the fact that ProfessorA works for University1 only holds during the period from 1994 to 2006. Time-sensitive data across multiple fields such as news, stocks, shopping carts, banking transactions, sensor networks, and social media networks are very common on the Web. The capability of representing time and space dimensions is essential in many real world systems and applications, especially in the systems extracting knowledge from the Web such as Knowledge Vault \cite{dong2014knowledge} recently created by Google, and YAGO2S \cite{hoffart2012yago2}. Other smart applications that use sensors, such as real-time remote monitoring of patients diagnosed with congestive heart failure, can benefit from temporal information \cite{hristoskova2014ontology}. Temporal and spatial databases have been researched over decades \cite{guting1994introduction,pelekis2004literature,snodgrass1986temporal} and supported by various commercial database systems. Enabling the representation of time and space in which RDF statements hold true is crucial to the development of various Semantic Web applications addressing such real world requirements of spatio-temporal content.


\textbf{Provenance}. 
Provenance provides information about the origin of an artifact or a piece of data. It involves entities, activities, and people/software/agents in the process of producing a piece of data. When it comes to Semantic Web, the provenance of an RDF assertion becomes important for integrating data from multiple sources. Data provenance can also be used for forming assessments about data quality, reliability, and trustworthiness. For example, is an assertion verified by a researcher more significant than one being reported in the literature?  
Tracking the lineage of data is essential in scientific workflows because it allows reproducibility. Reproducible datasets with explicit provenance information are usually considered more reliable, and hence, encouraged for sharing and publishing to the Web. Provenance is studied in various applications across different domains like biomedicine \cite{bodenreider2009provenance,wild2012systems,zhao2003annotating,zhao2004semantically}, geoscience \cite{di2013geoscience,yuan2013linked}, and others \cite{shu2012modelling}.


As metadata about triples is very common in modeling data, we believe that many Semantic Web datasets such as PubChem \cite{bolton2008pubchem}, Bio2RDF \cite{belleau2008bio2rdf}, and DisGeNET \cite{pinero2015disgenet} will benefit from a concensus on a standard method for representing triples about triples recommended by the W3C. However, such a concensus has yet to be reached.

Several attempts have been made in recent years to address the problem of representing metadata (e.g., provenance, time, and space) about RDF triples. \cite{callahan2013ovopub,groth2010anatomy,Nguyen:2014:DLR:2566486.2567973,sahoounified}. These approaches can be classified into three main groups: triple (reification \cite{hayes2004rdf}, singleton property\cite{Nguyen:2014:DLR:2566486.2567973}), quadruple (named graph \cite{carroll2005named}), and quintuple (RDF$^+$ \cite{schueler2008querying}). Among these approaches, the RDF reification, named graph, and RDF$^+$  have been studied in depth, but have failed to emerge as a standard for representing metadata about triples. While these approaches are intuitive, they all have certain limitations. 

An RDF triple is reified by creating an instance of RDF Statement class and pointing this instance to the subject, predicate, and object of the RDF triple. 
The reification of a triple does not entail the triple, and is not entailed by it \cite{hayes2004rdf}. As the presence of the reified triple is not bound to the triple itself, it is not possible to reason about the metadata of the triple through the reified triple. 

The named graph \cite{carroll2005named} originally introduced a fourth element to represent the provenance of the graph. Later, it was generalized to denote a set of RDF triples. A named graph can thus be used for annotating a triple and can serve as a unique identifier for this triple, which is not intuitive and represents a departure from the original intended use. 

The RDF$^+$ approach \cite{schueler2008querying} creates an internal identifier for each triple and uses it for asserting metadata about this triple. This approach defines abstract syntax and formal semantics for RDF$^+$ statements, creates mappings from RDF to RDF$^+$ and vice versa, and extends SPARQL to support RDF$^+$ statements. Therefore, it is incompatible with the Semantic Web standards.

Recently, the singleton property (SP) approach \cite{Nguyen:2014:DLR:2566486.2567973} has been proposed to address some of these outstanding problems. It creates one singleton property instance for each context to uniquely identify the triple, then uses this property instance to assert the metadata for this triple. For example, if ProfessorA works for University1 from 1994 to 2006, then we create a singleton property instance \verb|worksFor#1| and assert the temporal information for this property instance. Table \ref{singleton-example} displays how the motivating example is represented in RDF using the singleton property approach.

\begin{table}
\centering
\caption{Using the singleton property approach to represent the motivating example \label{singleton-example}}
\begin{tabular}{l @{\hspace{1em}}l @{\hspace{1em}}l @{\hspace{1em}}l} \hline \noalign{\smallskip}
Triple & Subject & Predicate & Object \\ \hline \noalign{\smallskip}
$T_0$ & ProfessorA & worksFor\#1& University1 \\ 
$T_1$ & worksFor\#1 & singletonPropertyOf & worksFor\\ 
$T_2$ & worksFor\#1 & from & 1994\\ 
$T_3$ & worksFor\#1 & to & 2006\\ 
$T_4$ & ProfessorA & worksFor\#2 & University2 \\ 
$T_5$ & worksFor\#2 & singletonPropertyOf & worksFor\\ 
$T_6$ & worksFor\#2 & from & 2007\\ 
$T_7$ & StudentB & hasAdvisor\#3 & ProfessorA \\ 
$T_8$ & hasAdvisor\#3 & singletonPropertyOf & hasAdvisor\\ 
$T_9$ & hasAdvisor\#3 & from & 2009\\ 
$M_7$ & worksFor & subPropertyOf & memberOf\\ \noalign{\smallskip}\hline\noalign{\smallskip}\end{tabular}
\end{table}

Unlike RDF$^+$, the SP approach represents metadata about triples in the form of triples, making it directly compatible with existing Semantic Web standards such as RDF and SPARQL. Furthermore, the current model-theoretic semantics can also be extended to include the semantics of singleton properties. 
This approach seems to be a potential candidate for representing metadata about triples. 
However, one inherent characteristic of this approach is that it generates a large property hierarchy in the schema because it relies on contextualized predicates, i.e., predicates used only once (``singleton'') in a specific context. This large property hierarchy may raise some important questions about its handling by the reasoners and query engines. 
\begin{enumerate}
\item How are the singleton triples related to the data triples they represent? 
If the singleton triples describe temporal information about a data triple, then how does a reasoner infer that data triple from the singleton triples? For example, the set ($T_0$, $T_1$, $T_2$, $T_3$) describes data triple $M_0$. How does one infer $M_0$ from ($T_0$, $T_1$, $T_2$, $T_3$)?
Similarly, if the set of data triples serves to infer a new data triple, can this data triple also be inferred by the singleton triples describing these data triples? In the motivating example at hand, the two triples $M_0$ and $M_7$ serve to infer a new data triple $M_8$: \verb|ProfessorA memberOf University1| according to the rule rdfs7 \cite{hayesrdf}. Can this data triple $M_8$ also be inferred by applying the rule rdfs7 to the triple $M_7$ and the singleton triples ($T_0$, $T_1$, $T_2$, $T_3$) describing the data triple $M_0$? 
\item If we load the singleton property triples in Table \ref{singleton-example} into a SPARQL endpoint, would the SPARQL engine be able to find the answers for the query with a mix of regular data triple patterns and singleton triple patterns as follows?\\
\verb|SELECT ?professor|\\
\verb|WHERE {?professor worksFor University2 .|\\
\verb|?student ?sp ?professor . |\\
\verb|?sp singletonPropertyOf hasAdvisor .|\\
\verb|?sp since "2009" .}|
\item Does the large singleton property hierarchy affect the performance of the reasoning and query answering tasks?
\end{enumerate}

We believe that these questions are critical and should be addressed for representing RDF statements about statements. It is also beneficial for data publishers and developers in the Semantic Web community to consider adopting this approach given its benefits compared to competing proposals. 
As the reasoning aspect was not tackled in our previous paper \cite{Nguyen:2014:DLR:2566486.2567973}, our goal here is to address these reasoning-related questions and perform a comprehensive evaluation. 









Our contributions in this paper include:

\begin{itemize}
\item A simple mechanism to enable existing reasoners to recognize the singleton triples and perform relevant reasoning tasks (Section \ref{mechanism}). We discuss the implementation of this mechanism and demonstrate it in the two reasoners, Oracle and Jena (Section \ref{implementation}). 
\item The benchmark LUBM-SP extended from LUBM to include singleton properties for representing temporal information (Section \ref{benchmark}).
\item Evaluation of the effect of singleton property triples on reasoning and query answering tasks (Section \ref{evaluation}).
\end{itemize}
We discuss related work in Section \ref{related}, future work in Section \ref{future}, and conclude with Section \ref{conclude}.

The data generator for the LUBM-SP datasets, together with its queries and the Jena package that are used in our evaluation are open source and available at \url{http://wiki.knoesis.org/index.php/Singleton_Property_Reasoning}.


\section{Reasoning Mechanism with Singleton Property}
\label{mechanism}
We start with a brief review of the singleton property \cite{Nguyen:2014:DLR:2566486.2567973}.
\subsection{Background}

A singleton property is defined as a property instance identifying the unique relationship between the subject and the object. For example, in order to represent the fact that ProfessorA worked for two universities at different times, we create two singleton property instances \verb|worksFor#1| and \verb|worksFor#2|, i.e., two instances of the property \verb|worksFor|. 
In general, given any triple $t$: ($s$, $p$, $o$) described by the contextual information represented by the pairs of meta property and meta value ($m^p_j$, $m^v_j$) with $j \in 1..n$, we create the following RDF representation summarized in Table \ref{singleton-graph-structure}. 
\begin{table}
\centering
\caption{Singleton graph structure asserting meta knowledge for data triple ($s$,$p$,$o$)\label{singleton-graph-structure}}
\begin{tabular}{l@{\hspace{1em}}l @{\hspace{1em}}l @{\hspace{1em}}l}  \noalign{\smallskip}
{\bf Triple} &{\bf Subject} & {\bf Predicate} & {\bf Object} \\ \noalign{\smallskip}
(1) & $p_i$& singletonPropertyOf & $p$ \\
(2) & $s$ & $p_i$ & $o$\\
(3) & $p_i$& $m^p_j$ & $m^v_j$\\\noalign{\smallskip}
\end{tabular}
\end{table}
First, we create a singleton property instance $p_i$ of the generic property $p$ (1), which we use to represent the unique relationship between $s$ and $o$ (2), and finally we assert the contextual information using this property instance $p_i$ (3).

The extension of the singleton property $p_i$ is the unique pair of subject and object ($s$, $o$), while the extension of the generic property $p$ is the set of all subject and object pairs associated with the singleton properties derived from it. In our example, the extension of the singleton property \verb|worksFor#1| is the pair (\verb|ProfessorA|, \verb|University1|), while the extension of the property \verb|worksFor| is the set of two pairs \{(\verb|ProfessorA|, \verb|University1|), (\verb|ProfessorA|, \verb|University2|)\}.

The property \verb|rdf:singletonPropertyOf| is defined in the RDF vocabulary by the formal semantics for singleton properties described in \cite{Nguyen:2014:DLR:2566486.2567973}, together with the list of primitive triples as follows: 

\begin{tabular}{ @{\hspace{1em}}l @{\hspace{1em}}l @{\hspace{1em}}l} \noalign{\smallskip}
rdf:singletonPropertyOf & rdf:type & rdf:Property \\ 
rdf:singletonPropertyOf & rdf:type & rdf:Resource \\ 
rdf:singletonPropertyOf & rdfs:domain & rdf:SingletonProperty \\ 
rdf:singletonPropertyOf & rdfs:range & rdf:Property\\ 
rdf:SingletonProperty & rdf:type & rdfs:Class \\ 
rdf:SingletonProperty & rdfs:subClassOf & rdf:Property .\\ 
\end{tabular}

Next, we present the simple mechanism to enable reasoning on singleton property triples.

\subsection{Primitive Axiom for SingletonPropertyOf}

First, let us have a closer look at the nature of the \verb|singletonPropertyOf| property.  
A generic property is more generic than all of its sub-properties. A generic property is also more generic than its singleton properties. For instance, the generic property \verb|worksFor| is more generic than the singleton property \verb|worksFor#1|. So one may ask if 
the \verb|singletonPropertyOf| property is equivalent to the property \verb|subPropertyOf|. However, the extension of a singleton property, e.g. \verb|worksFor#1|, has only one element (\verb|ProfessorA|, \verb|University1|). Therefore, the singleton property not only carries the semantics of its generic property (as sub-properties do), but it also carries the semantics of singleton-ness through its extension. Therefore, \verb|singletonPropertyOf| is not equivalent to \verb|subPropertyOf|.

If the two properties \verb|singletonPropertyOf| and \verb|subPropertyOf| are not equivalent, then what could be the relationship between them? As singleton property can be considered as a specialization, or sub property of a generic property, that suggests the new relationship as follows:

\begin{tabular}{l@{\hspace{1em}}l @{\hspace{1em}}l @{\hspace{1em}}l}  \noalign{\smallskip}
{\bf Triple} &{\bf Subject} & {\bf Predicate} & {\bf Object} \\ \noalign{\smallskip}
(4) & singletonPropertyOf & subPropertyOf & subPropertyOf \\
\end{tabular}

From the new triple (4) and the three triples (1), (2), and (3) in Table \ref{singleton-graph-structure}, we next show how the data triple can be derived. 

\subsection{Deriving Data Triples}
Back to the example from Section \ref{motivating}, how do we derive the triple \\$M_0$: \verb|ProfessorA worksFor University1| \\from the example set of singleton property triples?
Or, more generally, how do we derive the triple ($s$, $p$, $o$) from the set of triples (1), (2), (3), and (4)?
In practice, the data triples can be derived from singleton property triples in two passes.

In the first pass, we consider the two triples

\begin{tabular}{l @{\hspace{1em}}l @{\hspace{1em}}l @{\hspace{1em}}l} \noalign{\smallskip}
(1) & $p_i$& singletonPropertyOf & $p$ \\
(4) & singletonPropertyOf & subPropertyOf & subPropertyOf \\\noalign{\smallskip}
\end{tabular}

to derive a new triple (5) according to the rule rdfs7.

\begin{tabular}{l @{\hspace{1em}}l @{\hspace{1em}}l @{\hspace{1em}}l} \noalign{\smallskip}
(5) & $p_i$& subPropertyOf & $p$ \\\noalign{\smallskip}
\end{tabular}

In the second pass, we combine the two triples 

\begin{tabular}{l @{\hspace{1em}}l @{\hspace{1em}}l @{\hspace{1em}}l} \noalign{\smallskip}
(2) & $s$ & $p_i$ & $o$\\
(5) & $p_i$& subPropertyOf & $p$ \\\noalign{\smallskip}
\end{tabular}

to derive the data triple (6) also using rule rdfs7.

\begin{tabular}{l @{\hspace{1em}}l @{\hspace{1em}}l @{\hspace{1em}}l} \noalign{\smallskip}
(6) & $s$ & $p$ & $o$\\\noalign{\smallskip}
\end{tabular}

Therefore, after applying rule rdfs7 twice to the singleton property triples and the newly proposed axiom (4), we obtain the data triple (6): $s$ $p$ $o$. 

In our example, by applying the rule rdfs7 to the triples $T_1$ and (4), we get \\$T_{10}$: \verb|worksFor#1 subPropertyOf worksFor|. Then, we apply the rule rdfs7 again to the triple $T_{10}$ and $T_0$, to obtain the triple \\$M_0$: \verb|ProfessorA worksFor University1|.

Furthermore, one can also infer the triple $M_8$ from the subPropertyOf rule.

\begin{tabular}{l @{\hspace{1em}}l @{\hspace{1em}}l @{\hspace{1em}}l} \noalign{\smallskip}
$M_0$: & ProfessorA & worksFor & University1\\
$M_7$: & worksFor & subPropertyOf & memberOf\\
$M_8$: & ProfessorA & memberOf & University1\\\noalign{\smallskip}
\end{tabular}

Note that, as the data triple (6) is inferred from the singleton triples with certain contextual constraints, these contraints are also applicable to the data triple. In our example, the fact that \verb|ProfessorA worksFor University1| only holds during the time from 1994 to 2006, as it is inferred from the singleton property triples that express the duration constraint. How to further enforce the contraints on the data triple when it is involved in other deduction rules is another complex reasoning problem and is beyond the scope of this paper.
 










\subsection{Querying regular data pattern on singleton triples}
A dataset with metadata about the assertions would enable data consumers to represent and reason about the data quality or trust. However, not every query requires metadata about triples. We can think of the metadata about triples as optional information to query. In practice, a query may use only regular triple patterns, or only singleton triple patterns, or both. For example, one can query the list of universities which ProfessorA works for (regardless of time information) as follows:

\begin{tabular}{l @{\hspace{1em}}} \noalign{\smallskip}
\verb|SELECT ?university|\\ \noalign{\smallskip}
\verb|WHERE { ProfessorA worksFor ?university }|\\\noalign{\smallskip}
\end{tabular}

This query contains only one regular data triple pattern. Another query example specifying both types of triple patterns is the following query:

\begin{tabular}{l @{\hspace{1em}}} \noalign{\smallskip}
\verb|SELECT ?professor |\\
\verb|WHERE {|\\
\verb|?professor worksFor University2 .|\\
\verb|?student ?sp ?professor . |\\
\verb|?sp singletonPropertyOf hasAdvisor .|\\
\verb|?sp since "2009" . }|\\\noalign{\smallskip}
\end{tabular}

It asks for the list of professors who work for University2 and have student advisees since 2009. In both examples, a regular data triple pattern is specified in the queries. How would such queries be handled?

We consider two possibilities here. 

{\bf Dataset without data triples}. If the dataset contains only the singleton triples, the SPARQL queries like the above example may return an empty result set because the data triples are not present. In order to get the results, two conditions must be met. First, the query must be executed in an RDFS inference-enabled query engine that is able to compute the RDFS closure. Our evaluation confirms that Jena and Oracle support this feature. We have not thoroughly evaluated other query engines. Second, the axiom \\\verb|rdf:singletonPropertyOf rdfs:subPropertyOf rdfs:subPropertyOf| must be added to the schema to make the query reasoner precompute the inferred data triples from the singleton triples using RDFS rules. Here we explain how we implement this approach in Jena and Oracle. In Oracle, after loading the singleton property datasets into the RDF model, one must insert the singleton property axiom into this RDF model before creating any RDFS entailment rule index on this model. By default, the Oracle inference engine will run until the RDFS closure is reached. In Jena, we can simply create an in-memory OntModel with the spec RDFS\_MEM\_RDFS\_RULE. The RDFS closure will include all data triples inferred from the singleton property triples.
We will discuss details about where/how to add the singleton property axiom in Section \ref{implementation}.

{\bf Dataset with data triples.} If the dataset includes the set of data triples and the set of singleton triples, answering the query with the data triple pattern is trivial to any query engine. However, the question here is whether we should include the data triples as part of the dataset.
On the one hand, including data triples will speed up the execution for queries with data triple patterns. On the other hand, as data triples can be inferred from the singleton property triples, these data triples become redundant and may increase the risk of becoming inconsistent. If, for any reason, a data triple is removed from the dataset while the corresponding singleton property triples representing this data triple still remain in the dataset, then an inconsistency may result. However, this is a faulty update in the presence of singleton properties, not a problem with reasoning with singleton properties. 

\section{Implementation}
\label{implementation}

We have presented a simple mechanism for adding the singleton property axioms in order to entail the data triple represented by the singleton property triples. The singleton property axioms can be incorporated at any of the following levels.

{\bf Application.} One can add singleton property axioms into the application schema and perform reasoning tasks using reasoners that compute the RDFS closure, such as Jena or the Oracle inference engine \cite{wu2008implementing,wu2012advancing,zhou2013making}. Although this addition is straightforward, it requires the support of the tools being employed.

{\bf Reasoner tools.} Providing the support for the singleton property within a reasoner can also be done by the tool's creators. We have chosen Jena for our implementation because Jena is a popular open source Java API for RDF. We added the RDF \verb|SingletonProperty| class and \verb|singletonPropertyOf| property into the Java class RDF. We also added the primitive axioms and rules we propose for supporting singleton property reasoning to the rule bases. We have built a new package for jena-core
and used this package in our evaluation. Using this package, one does not need to further add the singleton property axioms into the application schema because it is already integrated into the rulebases. If there are singleton property triples in the knowledge base, the tool will recognize them and run the inference rules.



{\bf Language Specification.} Supporting singleton property as part of the RDF language specification will be the most effective level from the perspective of long-term standardization. We have seen enthusiasm from many Semantic Web practioners to implement this approach for their datasets. Others have been more reluctant because this approach is not standard. 
We are confident about the prospects of the wider adoption of this approach because the syntax and semantics of the singleton property are compatible with the current syntax and semantics of RDF, RDFS, and SPARQL. We have also explored the compatibility of the singleton property with OWL. This new capability can be incorporated into OWL by adding (i) the \verb|rdf:singletonPropertyOf| to the RDF vocabulary, and (ii) the singleton property axioms into the RDF schema. With these enhancements, the singleton property approach will be compatible with all Semantic Web standards, and will significantly enhance the quality and efficency of the meta reasoning.


\section{Knowledge Base Benchmark}
\label{benchmark}
In order to provide a benchmark for evaluating different aspects of the singleton property approach, as well as for future comparison with other approaches, we developed LUBM-SP by extending the LUBM data generator \cite{guo2005lubm}. 

\subsection{Data Generator}
In the LUBM knowledge bases, relationships are described among instances of 43 classes including Faculty, Student, Course, Publication, and Organization. Relationships asserted between two instances do not carry any contextual information, such as the duration or the source. LUBM has been extended to include the time dimension for RDF triples as in \cite{nguyen2013slubm} by stamping triples with dynamic or static temporal properties. However, the representation of this extended knowledge base is not in the standard form of RDF triples. Therefore, it cannot be used for evaluating tasks involving RDF or SPARQL.

While we also share the goal of enriching the LUBM knowledge base with the temporal information for the assertions, our approach differs from \cite{nguyen2013slubm} in that we represent the temporally-enriched LUBM using just RDF triples. We employ the singleton property approach for this purpose.
\begin{table}[h!]
\centering
\caption{The size of the LUBM-SP knowledge base is approximately twice the size of the original LUBM knowledge base\label{lubm-sizes}}
    \begin{tabular}{r @{\hspace{1em}}r @{\hspace{1em}}r @{\hspace{1em}}r@{\hspace{1em}}r}  \hline\noalign{\smallskip}
    \#Univs & LUBM        & LUBM:LUBM-SP   & LUBM-SP  &  Number of SPs \\\hline\noalign{\smallskip}
    50      & 6,654,562   & 1:2.0374 & 13,558,060 &  2,565,447 \\ \noalign{\smallskip}
    500     & 69,079,764  & 1:2.0035& 138,398,236  &  25,746,049 \\\noalign{\smallskip}
    1,000   & 138,278,374 & 1:2.0032 &  277,005,491 & 51,526,542 \\\noalign{\smallskip}
    5,000   &  690,885,862  &  1:2.0033 &  1,384,063,224  & 257,460,909 \\\hline\noalign{\smallskip}
    \end{tabular}
\end{table}

We modified the LUBM data generator to generate two different knowledge bases per run. The first one is the original LUBM knowledge base (KB). In the second one, for each triple of LUBM-SP that has temporal information attached, we replace this triple with the singleton property graph. All the information captured remains the same in the two knowledge bases. The only difference between them is representation, i.e., that one uses regular triples, while the other uses singleton property triples and temporal properties. For example, if the LUBM KB contains the data triple $M_0$, then the LUBM-SP KB represents it as a set of triples ($T_0$, $T_1$, $T_2$, and $T_3$) as provided in Table \ref{singleton-example}.

Table \ref{lubm-sizes} lists the number of triples in the LUBM and LUBM-SP knowledge bases for increasing numbers of universities (50, 500, 1000, and 5000). It also shows the corresponding number of singleton properties in the LUBM-SP KBs.
From the sample, the size of the LUBM-SP datasets is approximately twice the size of the LUBM datasets.
Out of 17 relations defined in LUBM, 10 relations are selected for singleton property representation. Five of them have two temporal properties and five of them have one temporal property.




\subsection{Queries}
We create one set of SPARQL queries for each of the two types of query: data triple pattern query and mixed triple pattern query.

\textbf{Data triple pattern queries.} LUBM comes with an original set of 14 SPARQL queries which contain only data triple patterns.
We use the same set of queries for evaluating the performance of both LUBM and LUBM-SP. It allows us to measure the extra time taken for inferring and querying singleton properties. From this set of queries, we create an equivalent set of SEMMATCH queries to be executed in Oracle. 
For every query, we tested that the result sets obtained from LUBM and LUBM-SP had the same size as expected.

\textbf{Mixed triple pattern queries.}  This query type has data triple patterns mixed with singleton property triple patterns and temporal properties. We create this set of queries by appending the singleton property triple pattern and temporal properties to the set of data triple pattern queries.



\section{Evaluation}
\label{evaluation}
This section presents our experiments on evaluating the impact of the singleton property representation on the reasoning tasks. In particular, we evaluate the performance of the reasoning tasks at the RDFS level on both in-memory (Jena) and storage-persistent (Oracle) reasoners. We measure the time taken for computing the RDFS closure and the number of inferred triples in each knowledge base. We measure the query answering performance by running the same query on both knowledge bases and recording the execution time.

\subsection{Experiment setup}

We used a single machine running Ubuntu 12.04 LTS with 256GB RAM and three hard drives. On this machine, we deployed the newly built jena-core package with singleton property support as explained in Section \ref{implementation}. We also installed Oracle 12c Release 1 on this machine. We used the Automatic Memory Management feature and spread tablespaces across different hard drives. This setting serves our evaluation purpose since we deploy the different knowledge bases onto the same system and database configuration, so that the difference in performance can be attributed to the impact of the singleton property representation.

Disclaimer: Oracle does not have official support for Ubuntu. Although we succeeded in installing Oracle 12c on Ubuntu 12.02 LTS and setting it up to work for the experiments, we did not attempt to tune the system to get the best performance out of Oracle database. 
Therefore, the results should be taken as an academic exercise rather than a performance benchmark for Oracle databases.

\subsection{Loading and Indexing Entailment Rules}

We created six knowledge bases (KBs). Each KB consists of either a LUBM or LUBM-SP dataset generated as described in Section \ref{benchmark}, the Univ-bench ontology and the singleton property axiom triple \\
\verb|rdf:singletonPropertyOf rdfs:subPropertyOf rdfs:subPropertyOf|.
In Oracle, we loaded each KB into one RDF model. 
For each model, we ran the PL/SQL procedure SEM\_APIS.CREAT\_ENTAILMENT to create an RDFS rulebase index with default options. Table \ref{oracle-inferencing} shows the number of inferred triples and the time taken for each index. 
 The number of inferred triples in the LUBM-SP KBs is larger than the one in the corresponding LUBM KBs. The time taken for creating the entailment index also increases in proportion with the number of inferred triples. 
We observed that the performance was bound to the heavy I/O accesses during the indexing processes. Therefore, increasing the number of disks may help increase the degree of parallelism, thereby improving the performance.

\begin{table}
    \centering
    \caption {Time taken for creating each RDFS entailment index in Oracle \label{oracle-inferencing}}
    \begin{tabular}{r@{\hspace{1em}}r@{\hspace{1em}}r@{\hspace{1em}}}\hline\noalign{\smallskip}
    {\bf KBs}          & {\bf \#Inferred Triples} & {\bf  Inferencing Time} \\ \hline \noalign{\smallskip}
    LUBM\_50     &  2,745,971             & 6.5 min               \\\noalign{\smallskip}
    LUBM-SP\_50  &  15,573,215            & 28 min               \\\noalign{\smallskip}
    LUBM\_500    &  27,510,369            & 57 min               \\\noalign{\smallskip}
    LUBM-SP\_500 &  102,050,743           & 6 h 41 min            \\\noalign{\smallskip}
    LUBM\_1000   &  55,069,401            & 1 h 58 min              \\\noalign{\smallskip}
    LUBM-SP\_1000 &  204,246,093           & 13 h 26 min            \\\hline\noalign{\smallskip}
    \end{tabular}
\end{table}

In Jena, we loaded each KB into an OntModel. The singleton property axiom need not be loaded into Jena models because it already has built-in support from our updated jena-core package. 
The average loading time after three runs is 35 seconds for LUBM\_50 and 79 seconds for LUBM-SP\_50. The average preparation time after three runs for initial processing and caching is 20 seconds for LUBM\_50 and 99 seconds for LUBM-SP\_50.

\subsection{Query Answering}




We used the set of queries described in Section \ref{benchmark}. Queries with mixed triple patterns are not applicable (N/A) to LUBM KBs because these KBs do not have singleton property triple patterns. We ran the set of queries on each model with a fresh database. Table \ref{query-answering} reports the average execution time after three runs for both Jena and Oracle. 

\begin{table}
    \centering
    \caption {Time taken per query and total time taken (excluding N/A and timed out) \label{query-answering}}
    \begin{tabular}{|r@{\hspace{0.5em}}|r@{\hspace{0.5em}}|r@{\hspace{0.5em}}|r@{\hspace{0.5em}}|r@{\hspace{0.5em}}|r@{\hspace{1em}}|}\hline
          & ~         &  \multicolumn{2}{ c| } {\bf Oracle}         & \multicolumn{2}{ c| } {\bf Jena (in-memory)}           \\ 
    \textbf{Query} &  \textbf{Resultset} &  \textbf{LUBM\_50} &  \textbf{LUBM-SP\_50} &  \textbf{LUBM\_50} &  \textbf{LUBM-SP\_50} \\  \hline \noalign{\smallskip}
    Q1     & 4         & 11.50    & 12.18       & 2.29    & 2.46           \\ \hline \noalign{\smallskip}
    Q2     & 130       & 14.44    & 107.77      & timeout        & timeout           \\ \hline \noalign{\smallskip}
    Q2-Mixed     & 130       & N/A    &  18.54     & N/A        & timeout           \\ \hline \noalign{\smallskip}
    Q3     & 6         & 5.90     & 6.25        & 6.74    & 6.97           \\ \hline \noalign{\smallskip}
    Q4     & 34        & 4.89     & 14.34       & 3.66    & 2.43           \\ \hline \noalign{\smallskip}
    Q5     & 719       & 12.84    & 10.60       & 23.95   & 37.57           \\ \hline \noalign{\smallskip}
    Q6     & 393,730    & 110.72   & 91.55       & 73.42   & 74.51           \\ \hline \noalign{\smallskip}
    Q7     & 59        & 5.74     & 3.15        & timeout       & timeout          \\ \hline \noalign{\smallskip}
    Q7-Mixed     & 59        & N/A     &   15.66      & N/A       &  timeout         \\ \hline \noalign{\smallskip}
    Q8     & 5,916      & 90.66    & 93.75       & timeout        & timeout           \\ \hline \noalign{\smallskip}
    Q9     & 6,538      & 133.02   & 118.73      & timeout        & timeout           \\ \hline \noalign{\smallskip}
    Q9-Mixed     & 6,538      & N/A   &  36.96     & N/A        &   timeout         \\ \hline \noalign{\smallskip}
    Q10    & 0         & 5.29     & 6.14        & 2.88    & 4.11           \\ \hline \noalign{\smallskip}
    Q11    & 0         & 2.44     & 1.78        & 0.11    & 0.12           \\ \hline \noalign{\smallskip}
    Q12    & 0         & 5.45     & 4.70        & 0.001    & 0.001           \\ \hline \noalign{\smallskip}
    Q13    & 0         & 3.99     & 3.75        & 1.8    & 2.11           \\ \hline \noalign{\smallskip}
    Q14    & 393,730    & 97.86    & 92.21       & 70.76   & 74.49           \\ \hline  \noalign{\smallskip}
    \multicolumn{2}{|l|}{\bf Total w/o mixed}      & {\bf 504.74}    & {\bf 566.90}       & {\bf 185.61}   & {\bf 204.78}           \\ \hline\noalign{\smallskip}    
    \end{tabular}
\end{table}
On average, the queries executed on the LUBM-SP KBs take 10\% (Jena) and 12\% (Oracle) longer than the ones executed on the LUBM KBs. This is expected due to the size of the KBs.
In Jena, queries running longer than two hours  were aborted. Queries 8 and 9, that are reported to have timed out in \cite{guo2005lubm} running Jena in-memory with LUBM\_1 (generated with 1 university), also did time out in our case.
Compared to Jena, these long queries are executed much faster in Oracle. The performance per query from LUBM and LUBM-SP KBs is not as consistent in Oracle as it is in Jena, because of the cache supported by Oracle. Subsequent queries may benefit from the cache loaded for prior queries. This is applicable to both original LUBM queries and the new mixed queries.

\textbf{Summary.}
Our evaluation shows the overall costs of reasoning tasks for KBs with and without singleton property representation. In general, the size of knowledge bases LUBM-SP with the singleton property representation is twice the size of knowledge bases LUBM with the data triple representation. The number of inferred triples determines the loading time and the execution time for creating rule indices. On average, the queries in LUBM-SP KBs take 10\% to 12\% longer than in LUBM KBs.

\section{Related Work}
\label{related}
Representing metadata about triples has received much attention and several approaches have been proposed. We can classify these approaches into three categories: triple-based (reification \cite{hayes2004rdf}, singleton property), quadruple-based (named graph \cite{carroll2005named}), and quintuple-based (RDF$^+$ \cite{schueler2008querying}) as explained in Section \ref{motivating}.
Our work is developed on top of the singleton property approach. One major advantage of the singleton property approach over reification is that the original data triple can be inferred by the singleton property triples describing it.

To the best of our knowledge, this is the first approach that allows the RDF data triple described by RDF triples to be inferred by an RDFS reasoner. This kind of reasoning differs from the stream reasoning \cite{nguyen2013slubm} where the temporal dimension is not represented directly with RDF. It also differs from the temporal RDF \cite{gutierrez2005temporal} where temporal reasoning is incorporated into RDF using reification. In contrast to these approaches, we demonstrated that temporal information can be incorporated into RDF through singleton properties to enable temporal reasoning.


\section{Future Work}
\label{future}



In the near future, we plan to address the following aspects:

\textbf{Performance optimization.} In the implementation and evaluation, our goal was to support singleton property reasoning. We have not attempted to optimize the reasoning performance. However, the singleton property hierarchy is usually large and better reasoning performance is always desirable for scalability. Evaluating this singleton property reasoning with more highly optimized RDF stores should also be done.

\textbf{Temporal RDF.} As the singleton property approach provides a mechanism for incorporating metadata such as temporal dimension into the RDF triples, work \cite{gutierrez2005temporal,hurtado2006reasoning} on reasoning about temporal RDF can also be revisited and reimplemented on top of singleton properties. Similarly, reasoning about other kinds of metadata describing triples should also be explored.


\textbf{Performance comparison}. We plan to compare the inference-enabled query performance with that of other approaches such as N-ary and Nanopub using life science KBs such as PubChem, Bio2RDF, and DisGeNET.

\section{Conclusion}
\label{conclude}
We have presented a simple mechanism for reasoning with RDF statements about statements. We have shown that this mechanism is in compliance with all Semantic Web standards and technically straightforward to implement. Our evaluation results also show acceptable overhead in the reasoning tasks involving the knowledge bases enriched with metadata such as time, location, and provenance for assertions, especially considering the conceptual simplicity and theoretical soundness.


\bibliographystyle{abbrv}
\bibliography{../../bibtex/all,../../bibtex/knowledge}

\end{document}